*Explorations in an English Poetry Corpus: A Neurocognitive Poetics Perspective*

Arthur M. Jacobs[1,2,3]

Author Note

1) Department of Experimental and Neurocognitive Psychology, Freie Universität Berlin,
Germany

2) Dahlem Institute for Neuroimaging of Emotion (D.I.N.E.), Berlin, Germany

3) Center for Cognitive Neuroscience Berlin (CCNB), Berlin, Germany

Correspondence: Arthur M. Jacobs

Department of Experimental and Neurocognitive Psychology, Freie Universität Berlin,
Habelschwerdter Allee 45 , D-14195 Berlin, Germany.

Email: ajacobs@zedat.fu-berlin.de

**Abstract**

This paper describes a corpus of about 3000 English literary texts with about 250 million words extracted from the Gutenberg project that span a range of genres from both fiction and non-fiction written by more than 130 authors (e.g., Darwin, Dickens, Shakespeare). Quantitative Narrative Analysis (QNA) is used to explore a cleaned subcorpus, the *Gutenberg English Poetry Corpus* (GEPC) which comprises over 100 poetic texts with around 2 million words from about 50 authors (e.g., Keats, Joyce, Wordsworth). Some exemplary QNA studies show author similarities based on latent semantic analysis, significant topics for each author or various text-analytic metrics for George Eliot's poem 'How Lisa Loved the King' and James Joyce's 'Chamber Music', concerning e.g. lexical diversity or sentiment analysis. The GEPC is particularly suited for research in Digital Humanities, Natural Language Processing or Neurocognitive Poetics, e.g. as training and test corpus, or for stimulus development and control.

## Introduction

In his 'The psycho-biology of language', Zipf (1932) introduced the law of linguistic change claiming that as the frequency of phonemes or of linguistic forms increases, their magnitude decreases. Zipfs law elegantly expresses a tendency in languages to maintain an equilibrium between unit length and frequency suggesting an underlying law of economy. Thus, Zipf speculated that humans strive to maintain an emotional equilibrium between *variety* and *repetitiveness* of environmental factors and behavior and that a speakers discourse must represent a compromise between variety and repetitiveness adapted to the hearers tolerable limits of change in maintaining emotional equilibrium. In a way, Zipf not only was a precursor of contemporary natural language processing/NLP (e.g., Natural Language Tool Kit/NLTK; Bird, Klein, & Loper, 2009), Quantitative Narrative Analysis (QNA), Computational Linguistics or Digital Humanities, but also of Psycholinguistics and Empirical Studies of Literature, since he theorized about 'the hearers responses' to literature.

About 30 years later, when analysing Baudelaires poem 'Les chats', Jakobson and Lévi-Strauss (1962) counted text features like the number of nasals, dental fricatives, liquid phonemes or adjectives and homonymic rhymes in different parts of the sonnet (e.g., the 1$^{st}$ quatrain) to support their qualitative analyses and interpretation of e.g. oxymora that link stanzas, of the relation between the images of cats and women, or of the poem as an open system which progresses dynamically from the quatrain to the couplet. While their systematic structuralist pattern analysis of a poem starting with formal metric, phonological and syntactic features to prepare the final semantic analysis provoked a controversy among literary scholars, it also settled the ground for subsequent linguistic perspectives on the analysis (and reception) of literary texts called *cognitive poetics* (e.g., Leech, 1969; Stockwell, 2002; Tsur, 1983; Turner & Poeppel, 1983).

Today, technological progress has produced *culturomics*, i.e. computational analyses of huge text corpora (5.195.769 digitized books containing ~4% of all books ever published) enabling researchers to observe cultural trends and subject them to quantitative investigation (Michel et al., 2011). More particularly, *Digital Literary Studies* now 'propose systematic and technologically equipped methodologies in activities where, for centuries, intuition and intelligent handling had played a predominant role' (Ganascia, 2015; Moretti, 2005).

One promising application of these techniques is in the emerging field of *Neurocognitive*

*Poetics* which is characterized by neurocognitive (experimental) research on the reception of more natural and ecologically valid stimuli focusing on *literary materials*, e.g. excerpts from novels or poems (Jacobs, 2015a,b; Schrott & Jacobs, 2011; Willems & Jacobs, 2016). These present a number of theoretical and methodological challenges (Jacobs & Willems, 2017) regarding experimental designs for behavioral and neurocognitive studies which –on the stimulus side– can be tackled by using advanced techniques of NLP, QNA and machine learning (e.g., Jacobs, Hofmann, & Kinder, 2016; Jacobs & Kinder, 2017, 2018; Jacobs, Schuster, Xue, & Lüdtke, 2017; Mitchell, 1997; Pedregosa et al., 2011).

A particularly interesting challenge consists in finding or creating the optimal training corpus for empirical scientific studies of literature (Jacobs, 2015c), since standard corpora are not based on particularly literary texts. Recently, Bornet and Kaplan (2017) introduced a literary corpus of 35 French novels with over 5 million word tokens for a *named entity recognition* study, but in the fields of psycholinguistics and (Neuro-)Cognitive Poetics such specific corpora still are practically absent.

In this paper, I describe a novel literary corpus assembled from the digitized books part of project Gutenberg (https://web.eecs.umich.edu/~lahiri/gutenberg_dataset.html), augmented by a Shakespeare corpus (http://www.shakespeare-online.com/sonnets/sonnetintroduction.html; cf. Jacobs et al., 2017), henceforth called the Gutenberg Literary English Corpus/*GLEC*. The GLEC provides a collection of over 3000 English texts from the Gutenberg project spanning a wide range of genres, both fiction and non-fiction (novels, biographies, dramas, essays, short stories, novellas, tales, speeches and letters, science books, poetry; e.g., Austen, Bronte, Byron, Coleridge, Darwin, Dickens, Einstein, Eliot, Poe, Twain, Woolf, Wilde, Yeats) with about *12 million* sentences and *250 million* words.

**The GLEC and GEPC**

The GLEC, i.e. the original Gutenberg texts augmented by the Shakespeare corpus, contains over 900 novels, over 500 short stories, over 300 tales and stories for children, about 200 poem collections, poems and ballads and about 100 plays, as well as over 500 pieces of non-fiction, e.g. articles, essays, lectures, letters, speeches or (auto-)biographies. Except for the poetry collection subcorpus further explored in this paper and henceforth called the *Gutenberg English Poetry Corpus* (GEPC), these texts are not (yet) edited, shortened or cleaned yet.

For the present analyses, I cleaned (in large part manually) all 116 texts making up the GEPC, e.g. by deleting duplicate poems, prefaces, introductions, content tables and indices of first lines, postscripts, biographical and author notes, as well as footnotes[1] or line and page numbers, and by separating poems from plays or essays (e.g., in Yeats texts), so that only the poems themselves remain in the texts without any piece of prose. This was important to obtain a valid 'poetry-only' subcorpus and a valid *poetic language model* for comparison with poetic texts or text fragments, such as metaphors (Jacobs & Kinder, 2017, 2018). Without such cleaning, computation of any *ngram model*, for instance, would be distorted by the prose parts. For the same reason, I also deleted poems in other languages than English, e.g. Lord Byrons 'Sonetto di Vitorelli', PB Shelleys 'Buona Notte' or TS Eliots 'Dans le Restaurant'.

In a second step, I concatenated all poetic texts written by a specific author which yielded a collection of 47 compound texts by the following authors: Aldous Huxley, Alexander Pope, Ambrose Bierce, Andrew Lang, Bret Harte, Charles Dickens, Charles Kingsley, DH Lawrence, Edgar Allan Poe, Elizabeth Barrett Browning, Ezra Pound, GK Chesterton, George Eliot, Herman Melville, James Joyce, James Russell Lowell, John Dryden, John Keats, John Milton, Jonathan Swift, Leigh Hunt, Lewis Carroll, Lord Byron, Lord Tennyson, Louisa May Alcott, Oscar Wilde, PB Shelley, Ralph Waldo Emerson, Robert Browning, Robert Frost, Robert Louis Stevenson, Robert Southey, Rudyard Kipling, Samuel Taylor Coleridge, Shakespeare, Sir Arthur Conan Doyle, Sir Walter Scott, Sir William Schwenck Gilbert, TS Eliot, Thomas Hardy, Walt Whitman, Walter de la Mare, William Blake, William Butler Yeats, William Dean Howells, William Makepeace Thackeray, William Wordsworth. These 47 compound texts differ in a variety of surface and deep structure features, some of which are analyzed in the following sections. As can bee seen in Table A1 in the Appendix, text length also varies considerably across authors (exponential distribution with a median of 23000 words): the top three authors are Lord Byron (~ 210.000 words), PB Shelley (~165.000) and Wordsworth (~115.000), the 'flop' three are Alcott (<400), Pound (~1200) and Joyce (~1200). The majority of texts have less than 40.000 words. The entire GEPC comprises *1.808.160* words (tokens) and 41.857 types.

**Some exemplary analyses of the GEPC**

---

[1] For example, there were over 1000 footnotes in 'Lord Byrons Poetical Works Vol. 1' occupying a notable portion of the entire text.

Like other fields (e.g., Computational Linguistics or Digital Humanities), *Neurocognitive Poetics* (Jacobs, 2015a) uses text corpora for many purposes. An important one is the computation of a *language model*, usually based on trigrams (e.g., Jurafsky & Martin, 2016). Trigram probabilities are then used to compute variables such as *surprisal* which are reliable and valid predictors of a number of response measures collected in empirical research on reading and literature, e.g. reading time or brain wave amplitudes (Frank, 2013). However, it makes a big difference when trigram probabilities are computed on the basis of a nonliterary as compared to a literary or poetic training corpus (cf. Jacobs et al., 2017).

Other purposes are *similarity analyses*, which can be based on features extracted by latent semantic, topic or sentiment analyses (e.g., Deerwester et al., 1990; Jacobs et al., 2015; Schmidtke et al., 2014a; Turney & Littman, 2003). Such features can then be used to train classifiers for identifying authors, periods of origin or main motifs, as well as for predicting ratings and other response data of poetic texts (e.g. Jacobs et al., 2017; Jacobs & Kinder, 2017, 2018; Stamatatos, 2009; van Halteren et al., 2005).

**Similarity analyses**

As an example for a similarity analysis, Figure 1 shows a multidimensional scaling representation of the 47 texts of the GEPC based on latent semantic analysis (document-term-matrix/DTM analysis[2]). This analysis reveals e.g. that the 'Lake poets' (e.g., Coleridge, Woodsworth) cluster together, or that some poets like Pound or Joyce stand out from the rest.

Figure 1 here

Figure 2 shows a heatmap comparing the 20 most significant topics (as extracted by *Non-Negative Matrix Factorization/NMF*; Pedregosa et al., 2011) for the 47 texts (a list of the 20 most significant words per topic is given in the Appendix). The color code is proportional to the probability of a given topic, i.e. all 20 values per author add up to 1.

Figure 2 here

The data summarized in Figure 2 and the topic list (Appendix) reveal e.g. that the most important topic for Shakespeare's sonnets is topic #16 represented by the following 20 key word stems: 'natur spirit hath everi truth right hope think doth back find much faith art free round whole set drop'. In contrast, topic #2 appears to be the most important for Lord Tennyson (key word stems: 'king knight arthur round queen answer mine lancelot saw lord mother arm call name thine hall among child hath speak').

---

[2] The DTM was based on sklearn-CountVectorizer with minimum term frequency = 1; maximum term frequency = .95, NLTK stopwords; stemmer = SnowballStemmer (cf. Bird, Klein, & Loper, 2009; Pedregosa et al., 2011).

Moreover, the data of Figure 2 reveal that texts like those of Walt Whitman cover only four of the 20 topics (#1,3,4,17 have probabilities > 0), whereas other authors such as Charles Dickens cover a large range of topics (i.e., 15/20 with p>0). Such data can be used further in deeper analyses of generic poetic texts such as Shakespeare sonnets that look at topics important for aesthetic success (e.g., Simonton, 1990) or for evoking specific affective and aesthetic reader responses (Jacobs et al., 2017).

**Comparing word uniqueness and distinctiveness for two texts**

A third and last example for how to use the present GEPC concerns a more detailed comparative analysis for a subset of the 47 texts including surface and semantic features. This is done for two authors with shorter texts of comparable length: Blake vs. Dickens (4439 words vs 3758). We have recently provided an extensive comparative QNA of all 154 Shakespeare sonnets looking at both surface and deep semantic features. For example, we compared features such as poem or line *surprisal*, *syntactic simplicity*, *deep cohesion*, or *emotion and mood potential* (Jacobs et al., 2017). As an example for another interesting feature not considered in our previous study, here I will focus on word distinctiveness or *keyness*. In computing this feature I closely followed the procedure proposed in: DARIAH – Digital Research Infrastructure for the Arts and Humanities; https://de.dariah.eu/tatom/feature_selection.html. According to DARIAH's operationalisation, one way to consider words as distinctive is when they are found exclusively in texts associated with a single author (or group). For example, if Dickens uses the word 'squire' in the present GEPC and Blake never does, one can count 'squire' as distinctive or unique (in this comparative context). Vice versa, the word 'mother' is distinctive in this GEPC comparison, because Dickens never uses it.

Identifying unique words simply requires to calculate the average rate of word use across all texts for each author and then to look for cases where the average rate is zero for one author. Based on the DTMs for both texts this yielded the following results:

**Table 1. Five unique words with usage rates (1/1000) in Blake's and Dickens' poems of the GEPC**

| Author/words | SQUIRE | LUCI | MOTHER | FINE | LAMB |
|---|---|---|---|---|---|
| Blake | 0 | 0 | 7.9 | 0 | 7.2 |
| Dickens | 11.2 | 8.7 | 0 | 7.5 | 0 |

Another approach to measuring *keyness* is to compare the average rate at which authors use a word by calculating the difference between the rates. Using this measure I calculated the top five distinctive words in the Blake-Dickens comparison by dividing the difference in both authors' average rates by the average rate across all 47 authors.

**Table 2. Top five distinctive words (stems) with usage rates (1/1000) in Blake's and Dickens' poems of the GEPC**

| Author/words | dol' | outgleam | chalon | toor | vithin |
|---|---|---|---|---|---|
| Blake | 0 | 0 | 0 | 0 | 0 |
| Dickens | .83 | .41 | .41 | .83 | .41 |

Thus, appearing only once in the entire text, Dickens' word stem 'outgleam' in the line 'Behold outgleaming on the angry main!' appears to be distinctive, much as the other four word stems in Table 2.

A final quantitative comparison inspired by DARIAH's approach to determining word distinctiveness uses Bayesian group or author comparison. It involves estimating the belief about the observed word frequencies to differ significantly by using a probability distribution called the sampling model. This assumes the rates to come from two different normal distributions and the question to be answered is how confident one is that the means of the two normal distributions are different. The degree of confidence (i.e., a Bayesian probability), that the means are indeed different then is another probabilistic measure of distinctiveness.

Using a *Gibbs sampler* to get a distribution of posterior values for $\delta^3$, which is the variable estimating the belief about the difference in authors' word usage (for details, see https://de.dariah.eu/tatom/feature_selection.html.), I computed the probability that using the words 'squire' and 'fine' (both more characteristic of Dickens' poems than of Blake's) is likely to be zero.

**Table 3. Bayesian probability estimates (based on 2000 samples) for two distinctive words with usage rates (1/1000) in Blake's and Dickens' poems of the GEPC**

|                | SQUIRE | FINE |
|----------------|--------|------|
| p(delta<0)     | 0.23   | 0.09 |
| Blake average  | 0      | 0    |
| Dickens average| 11.2   | 7.5  |

According to this Bayesian analysis 'squire' appears more distinctive of Dickens' poetry than 'fine', but since both words do not produce a high probability of differing from zero, I would not put much belief in them being specifically characteristic of Dickens in the GEPC (although they are most distinctive in comparison to Blake, see Table 1). This Bayesian 'feature selection' method can be extended to every word occurring in a corpus producing a useful ordering of characteristic words (for details, see https://de.dariah.eu/tatom/feature_selection.html.).

**Comparing two individual poems**

The above analyses dealt with the entire GEPC or two poem collections, respectively. Next I focus on a more detailed –purely descriptive– comparison of two short individual texts from the GEPC that are far apart from each other (and the rest of the poems) in the similarity graph of Figure 1: George Eliot's poem 'How Lisa Loved the King' and James Joyce's 'Chamber Music'. I will give just a few illustrative statistics both for surface and deeper semantic features that are of potential use in Digital Humanities and Neurocognitive Poetics studies (for review on the latter, see Jacobs, 2015a; Jacobs et al., 2017).

---

[3] It represents half the difference between the population means for the distributions characterizing word rates in Blake and Dickens.

Two features that are often used as indicators of linguistic complexity, poetic quality or aesthetic success are *lexical diversity* –measured by the *type-token ratio*– and *adjective-verb quotient:* for example, 'better' Shakespeare sonnets are distinguished by a higher type-token ratio, more unique words, and a higher adjective-verb quotient (e.g., Simonton, 1989). The number of types can also be considered a co-estimate of the size of an authors' (active) mental lexicon and vocabulary profile. As can be seen in columns 2 and 3 of Table 4, both poems descriptively do not differ much on these features.

**Table 4. Some exemplary statistics for two poems**

| Author | Nbr. of word tokens / types / hapaxes / Type-token ratio (lexical diversity) / type% = types/41857 | Nbr. of Nouns, Verbs, Adjectives / Adjective-Verb quotient | Most Freq. Nouns, Verbs, Adjectives | Most Freq. Bi- & Trigram Collocations | Mean Sonority Score | Mean positive and negative valence, and arousal / most positive, negative and arousing word |
|---|---|---|---|---|---|---|
| Eliot | 2702, 1467, 1014, .5, 8% | 1111, 686, 642. .93 | LOVE(19), LIFE(15),SOUL(12) love(7),see(5),live(3) little(13),high(9),good(9) | 'King Pedro'(4), 'day might'(2), 'death tell'(2), 'Six hundred years'(2), 'Hundred years ago'(2), ''T gentle Lisa'(1) | 5.19 | 1.01, 0.84, 2.01 happiness, shame, happiness |
| Joyce | 1221, 654, 447, .53, 3% | 507, 313, 270, .86 | LOVE(23), HEART(18),AIR(9) love(7),come(3),sleep(2) sweet(13),soft(9),sad(5) | 'true love' (4), 'long hair'(3), 'pretty air'(3), 'combing long hair'(2), 'would sweet bosom '(2), 'singing merry air' (2) | 5.26 | 1.02, 0.85, 2.03 happiness, sadness, happiness |

Looking at the three most frequent nouns, verbs and adjectives, as well as significant bi- and trigram collocations in columns 4 and 5, the key words suggest that both poems have much to say about one of three favorite poetry motifs, i.e. *love*. This is also evident from the two lexical dispersion plots in Figure 3 that show, among others, that 'love' appears well distributed across the entire poems, never letting the reader forget the poems' central motif.

Figure 3 here

Poetic language expertly plays with the sound-meaning nexus and our group has provided empirical evidence that sublexical phonological features play a role in (written) poetry reception (Aryani et al., 2013; 2016; Jacobs, 2015b,c; Jacobs et al., 2015, 2016b; Schmidtke et al., 2014b; Schrott & Jacobs, 2011; Ullrich et al., 2017). A sublexical phonological feature with poetic potential is the *sonority score* (Jacobs, 2017; Jacobs & Kinder, 2018; see Appendix A for details). It is based on the notion of sonority profile (cf. Clements, 1990; Stenneken et al., 2005) which rises maximally toward the peak and falls minimally towards the end, proceeding from left to right, for the universally preferred syllable type (Clements, 1990, p. 301). Through a process of more or less unconscious *phonological recoding* text sonority may play a role even in silent reading (Braun et al., 2009; Ziegler & Jacobs, 1995) and especially in reading poetic texts (Kraxenberger, 2017). Column six of Table 4 shows that the two poems differ little in their global sonority score. At a finer-grained level of individual lines or stanzas, sonority could still notably differ, however, and implicitly affect readers' affective-aesthetic evaluation (cf. Jacobs & Kinder, 2018)

An important task for QNA-based Neurocognitive Poetics studies is *sentiment analysis*, i.e. to estimate the emotional valence or mood potential of verbal materials (e.g., Jacobs et al., 2017). In principle, this is done with either of two methodological approaches: using word lists that provide values of word valence or arousal based on human rating data (e.g., Jacobs et al., 2015), or applying a method proposed by Turney and Littman (2003) based on associations of a target word with a set of *labels*, i.e. key words assumed to be prototypical for a certain affect or emotion. Following previous research (Westbury et al., 2014), I computed the lexical features *valence* and *arousal* according to a procedure described in Appendix B.

The mean values in the rightmost column of Table 4 indicate that at this global level both poems practically do not differ on any of these three affective features. This can be visualized for the entire poems by the 3D plots of the principal components extracted from the three variables for all words in the poems: descriptively, they appear very similar. All other things being equal, this suggests that e.g. human ratings of the global affective meaning of both poems should not differ significantly (cf. Aryani et al., 2016).

Figure 4 here

## Discussion

In this paper I have briefly described a relatively big corpus of English literary texts, the GLEC, for use in studies of Computational Linguistics, Digital Humanities or Neurocognitive Poetics. As a whole, the GLEC requires further processing (e.g, cleaning, regrouping according to subgenres etc.) before it can be used as a training and/or test corpus for future studies. Using a smaller subcorpus already cleaned and consisting of 116 poetry collections, poems and ballads from 47 authors, i.e. the GEPC, I presented a few exemplary QNA studies in detail. In these explorations of the GEPC, I showed how to use similarity and topic analyses for comparing and grouping texts, several methods for identifying distinctive words, and procedures for quantifying important features that influence reader responses to literary texts, e.g. lexical diversity, sonority score, valence or arousal.

The GEPC could be applied to a variety of research questions such as authorship and period of origin classifications (cf. Stamatatos, 2009), the prediction of beauty ratings for poetic stimuli (e.g., Jacobs & Kinder, 2017, 2018) or the design of neuroimaging studies using literary stimuli (e.g., Bohrn et al., 2013; O'Sullivan et al., 2015). A limitation of the corpus lies in its texts being relatively 'old': due to copyright issues, the GLEC and GEPC contain only texts from 1623 to 1952, the majority of the GEPC stemming from the 19[th] century (Median = 1885). To what extent the GLEC or GEPC are useful training corpora for studies using more modern and contemporary literary text materials is an open empirical question to be adressed in future research. The successful application of the GLEC as a reliable language model (with a hit rate of 100%) for the computation of the surprisal values of 464 metaphors which also included contemporary ones (Katz et al., 1988) is encouraging in this respect (Jacobs & Kinder, 2018).

**Figure Captions**

**Figure 1.** Multidimensional scaling plot of text similarity for 47 authors from the GEPC based on latent semantic analysis (DTM comparisons).

**Figure 2.** Heat map of 20 main topics for the 47 texts of the GEPC based on Non-Negative Matrix Factorization/NMF

**Figure 3.** Lexical dispersion plots for the three most frequent nouns, verbs and adjectives in George Eliot's poem 'How Lisa Loved the King' and James Joyce's 'Chamber Music'. Note that the word 'love' is counted both as noun and verb here.

**Figures**

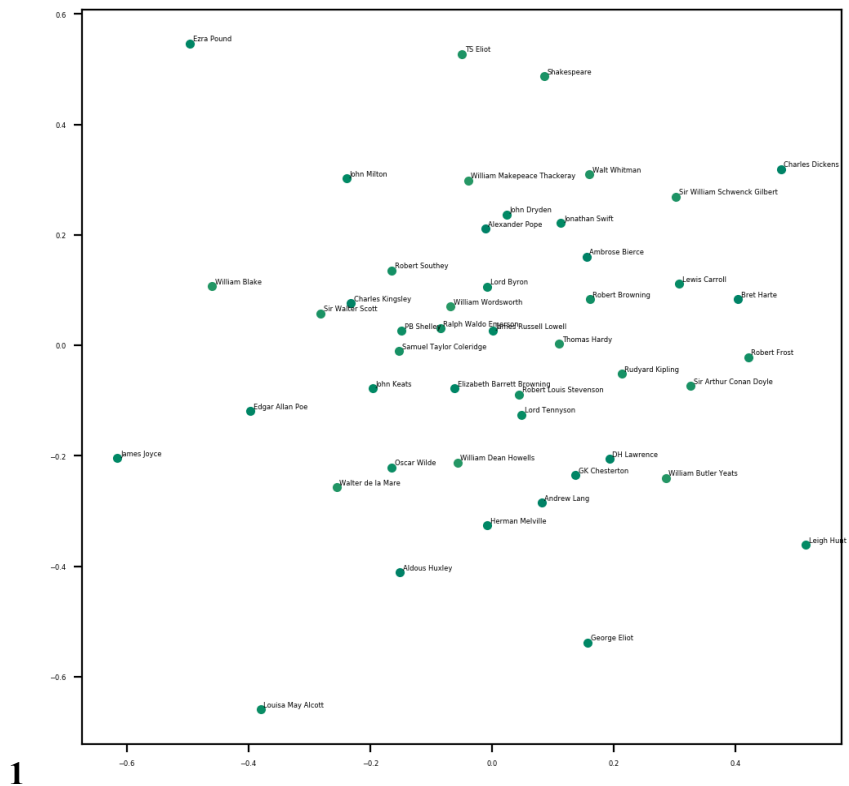

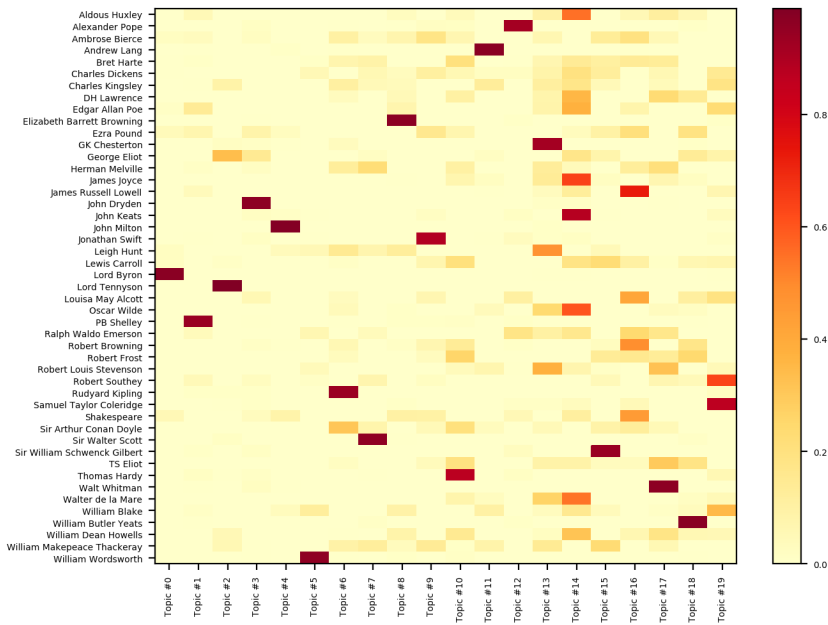

**2**

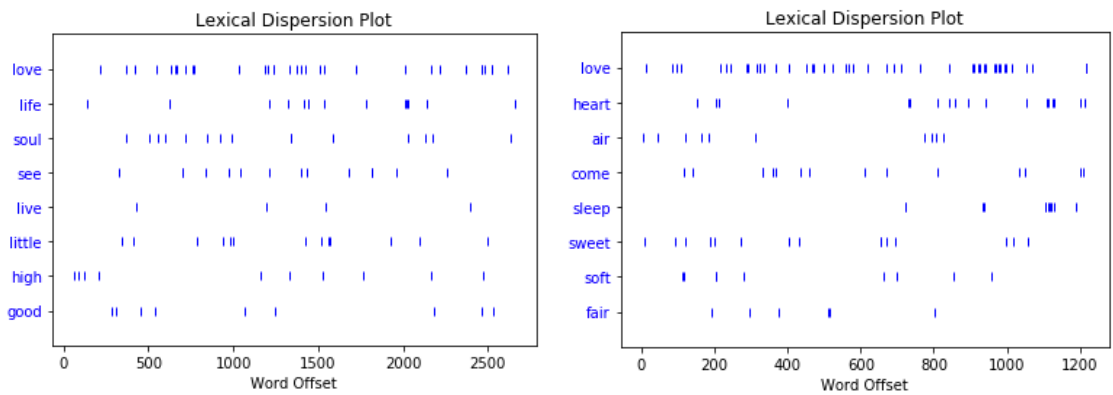

**3**

Appendix

**Table A1. List of authors with example texts in the GLEC and total text lengths in the GEPC**

| Author | Nbr. of Texts | Example Text in GLEC, year of publication | *GEPC Text, length (nbr. of words)* |
|---|---|---|---|
| 1. Abraham Lincoln | 16 | Lincoln's First Inaugural Address, 1861 | - |
| 2. Agatha Christie | 2 | The Secret Adversary, 1922 | - |
| 3. Albert Einstein | 2 | Relativity/ The Special and General Theory, 1916 | - |
| 4. Aldous Huxley | 3 | Crome Yellow, 1921 | *The Defeat of Youth and Other Poems, 4616* |
| 5. Alexander Pope | 3 | The Rape of the Lock and Other Poems, 1875 | *The Poetical Works, 82870* |
| 6. Alfred Russel Wallace | 5 | Is Mars Habitable?, 1907 | - |
| 7. Ambrose Bierce | 18 | A Cynic Looks at Life, 1912 | *Black Beetles in Amber, 23815* |
| 8. Andrew Lang | 60 | Historical Mysteries, 1904 | *A Collection of Poems, 46466* |
| 9. Anthony Trollope | 71 | The Eustace Diamonds, 1871 | - |
| 10. Arnold J. Toynbee | 1 | Turkey / A Past and a Future, 1917 | - |
| 11. Baronness Orczy | 16 | The Tangled Skein, 1907 | - |
| 12. Beatrix Potter | 1 | A Collection of Beatrix Potter Stories, 1902 | - |
| 13. Benjamin Disraeli | 17 | Vivian Grey, 1826 | - |
| 14. Benjamin Franklin | 4 | Autobiography of Benjamin Franklin, Version 4, 1791 | - |
| 15. Bertrand Russell | 8 | The Analysis of Mind, 1921 | - |
| 16. Bram Stoker | 6 | Dracula, 1897 | - |
| 17. Bret Harte | 58 | The Queen of the Pirate Isle, 1886 | *East and West, 6737* |
| 18. Charles Darwin | 20 | The Expression of Emotion in Man and Animals, 1859 | - |
| 19. Charles Dickens | 60 | Oliver Twist, 1837 | *The Poems and Verses, 3758* |
| 20. Charles Kingsley | 44 | True Words for Brave Men, 1884 | *Poems, 14391* |
| 21. Charlotte Bronte | 4 | Jane Eyre, 1847 | - |
| 22. DH Lawrence | 19 | Women in Love, 1920 | *Collected Poems, 19820* |
| 23. Edgar Allen Poe | 11 | The Masque of the Red Death, 1842 | *Complete Poetical Works, 8117* |
| 24. Edgar Rice Burroughs | 25 | Tarzan of the Apes, 1912 | - |
| 25. Edmund Burke | 15 | Burke's Speech on Conciliation with America, 1775 | - |
| 26. Edward P Oppenheim | 53 | The Zeppelin's Passenger, 1918 | - |
| 27. Elizabeth B Browning | | Sonnets from the Portuguese, 1850 | *The Poetical Works, 59404* |
| 28. Emily Bronte | 1 | Wuthering Heights, 1847 | - |

| | | | | |
|---|---|---|---|---|
| 29. | Ezra Pound | 2 | Certain Noble Plays of Japan, 1916 | *Hugh Selwyn Mauberley, 1181* |
| 30. | George A Henty | 89 | Under Drake's Flag, 1883 | - |
| 31. | George Bernard Shaw | 42 | Pygmalion, 1912 | - |
| 32. | George Eliot | 13 | Middlemarch, 1871 | *How Lisa Loved the King, 2702* |
| 33. | George Washington | 1 | State of the Union Addresses of George Washington, 1790 | - |
| 34. | GK Chesterton | 39 | The Wisdom of Father Brown, 1914 | *Complete Poems, 29867* |
| 35. | Hamlin Garland | 22 | Money Magic, 1907 | - |
| 36. | Harold Bindloss | 43 | Delilah of the Snows, 1907 | - |
| 37. | Harriet EB Stowe | 12 | Uncle Tom's Cabin, 1852 | - |
| 38. | Hector Hugh Munro | 7 | The Toys of Peace, 1919 | - |
| 39. | Henry David Thoreau | 9 | Walden, and On The Duty Of Civil Disobedience, 1854 | - |
| 40. | Henry James | 72 | The Golden Bowl, 1904 | - |
| 41. | Henry Rider Haggard | 52 | Love Eternal, 1918 | - |
| 42. | Herbert George Wells | 51 | The War of the Worlds, 1897 | - |
| 43. | Herbert Spencer | 4 | The Philosophy of Style, 1880 | - |
| 44. | Herman Melville | 16 | Moby Dick, 1851 | *Poems, 19088* |
| 45. | Howard Pyle | 11 | The Merry Adventures of Robin Hood, 1883 | - |
| 46. | Isaac Asimov | 1 | Youth, 1952 | - |
| 47. | Jack London | 48 | The Sea-Wolf, 1904 | - |
| 48. | Jacob Abbott | 47 | William the Conqueror, 1849 | - |
| 49. | James Bowker | 1 | Goblin Tales of Lancashire, 1878 | - |
| 50. | James F Cooper | 36 | The Last of the Mohicans, 1826 | - |
| 51. | James Joyce | 4 | Ulysses, 1922 | *Chamber Music, 1221* |
| 52. | James Matthew Barrie | 23 | Peter Pan, 1911 | - |
| 53. | James Otis (Kaler) | 27 | Dick in the Desert, 1893 | - |
| 54. | James Russell Lowell | 11 | Abraham Lincoln, 1890 | *The Complete Poetical Works, 45204* |
| 55. | Jane Austen | 8 | Emma, 1815 | - |
| 56. | Jerome K Jerome | 30 | Three men in a boat, 1898 | - |
| 57. | John Bunyan | 9 | The Holy War, 1682 | - |
| 58. | John Dryden | 13 | All for Love, 1678 | *The Poetical Works, 80667* |
| 59. | John Galsworthy | 40 | The Forsyte Saga, 1906-1921 | - |
| 60. | John Keats | 6 | Endymion, 1818 | *Poems, 36408* |
| 61. | John Locke | 3 | An Essay Concerning Humane Understanding, 1689 | - |
| 62. | John Maynard Keynes | 1 | The Economic Consequences of the Peace, 1919 | - |
| 63. | John Morley | 28 | On Compromise, 1874 | - |
| 64. | John Ruskin | 42 | A Joy For Ever, 1885 | - |
| 65. | John Stuart Mill | 11 | Utilitarianism, 1861 | - |
| 66. | Jonathan Swift | 15 | Gullivers Travels, 1726 | *The poems, 85834* |
| 67. | Joseph Conrad | 34 | Lord Jim, 1899 | - |
| 68. | Leigh Hunt | 3 | Stories from the Italian Poets/ With Lives of the Writers, 1835 | *Captain Sword and Captain Pen, 2260* |
| 69. | Lewis Carroll | 14 | Symbolic Logic, 1896 | *Poems, 15505* |
| 70. | Lord Byron | 12 | Fugitive Pieces, 1806 | *Poetical Works, 207977* |
| 71. | Lord Tennyson | 10 | Lady Clara Vere de Vere, 1842 | *The Poems, 105650* |
| 72. | Louisa May Alcott | 34 | Little Women, 1869 | *Three Unpublished Poems, 386* |
| 73. | Lucy M Montgomery | 17 | Anne Of Green Gables, 1908 | - |
| 74. | Lyman Frank Baum | 42 | The Wonderful Wizard of Oz, 1900 | - |
| 75. | Mark Twain | 46 | The Adventures of Tom Sawyer, 1876 | - |
| 76. | Mary Shelley | 5 | Frankenstein, 1818 | - |
| 77. | Michael Faraday | 2 | Experimental Researches in Electricity, 1839 | - |
| 78. | Mary Stewart Daggett | 2 | Mariposilla, 1895 | - |
| 79. | Nathaniel Hawthorne | 88 | The Scarlet Letter, 1850 | - |
| 80. | O Henry | 14 | The Gift of the Magi, 1905 | - |
| 81. | Oscar Wilde | 25 | The Picture of Dorian Gray, 1890 | *Poems, 22089* |
| 82. | PB Shelley | 7 | Adonais, 1821 | *The Complete Poetical Works, 165242* |
| 83. | PG Wodehouse | 35 | A Damsel in Distress, 1919 | - |
| 84. | Percival Lowell | 2 | The Soul of the Far East, 1896 | - |
| 85. | Philip Kindred Dick | 11 | Mr. Spaceship, 1953 | - |
| 86. | R M Ballantyne | 88 | The Red Eric, 1863 | - |
| 87. | Rafael Sabatini | 17 | Scaramouche, 1921 | - |
| 88. | Ralph Waldo Emerson | 7 | Nature, 1836 | *Poems, 29446* |
| 89. | Richard B Sheridan | 5 | Scarborough and the Critic, 1751 | - |
| 90. | Robert Browning | 7 | Men and Women, 1855 | *Poems, 35732* |
| 91. | Robert Frost | | A Boy's will, 1913 | *Poems, 15518* |
| 92. | Robert Hooke | 1 | Micrographia, 1665 | - |
| 93. | Robert L Stevenson | 79 | A Childs Garden of Verses, 1885 | *Poems, 33755* |
| 94. | Robert Southey | 3 | The Life of Horatio Lord Nelson, 1798 | *Poems, 23857* |
| 95. | Rudyard Kipling | 42 | The Jungle Book, 1894 | *Poems, 64137* |
| 96. | Samuel T Coleridge | 13 | The Rime Of The Ancient Mariner, 1798 | *The Complete Poetical Works, 51983* |
| 97. | Sinclair Lewis | 7 | Babbitt, 1922 | - |
| 98. | Sir Arthur Conan Doyle | 57 | The Adventures of Sherlock Holmes, 1892 | *Poems, 14386* |
| 99. | Sir Francis Galton | 3 | Inquiries into Human Faculty and Its Development, 1883 | - |
| 100. | Sir Humphry Davy | 1 | Consolations in Travel, 1830 | - |
| 101. | Sir Isaac Newton | 3 | Opticks, 1704 | - |
| 102. | Sir Joseph Dalton Hooker | 1 | Himalayan Journals, 1854 | - |
| 103. | Sir Richard Francis Burton | 11 | The Land of Midian, 1877 | - |
| 104. | Sir Walter Scott | 35 | Ivanhoe, 1820 | *Poems, 46846* |
| 105. | Sir Winston Churchill | 4 | The River War, 1899 | - |
| 106. | Sir William Schwenck Gilbert | 5 | Songs of a Savoyard, 1890 | *Poems, 31138* |
| 107. | Stephen Leacock | 15 | Frenzied Fiction, 1917 | - |
| 108. | TS Eliot | 4 | The Waste Land, 1922 | *Poems, 4661* |

| 109. Thomas Carlyle | 32 | History of Friedrich II of Prussia, 1895 | - |
| 110. Thomas Crofton Croker | 1 | A Walk from London to Fulham, 1813 | - |
| 111. Thomas Hardy | 26 | Tess of the d'Urbervilles, 1891 | *Poems, 62756* |
| 112. Thomas Henry Huxley | 44 | Darwinian Essays, 1893 | - |
| 113. Thomas Robert Malthus | 4 | An Essay on the Principle of Population, 1798 | - |
| 114. Thornton Waldo Burgess | 31 | Mrs. Peter Rabbit, 1902 | - |
| 115. Ulysses Grant | 3 | State of the Union Addresses, 1875 | - |
| 116. Virginia Woolf | 4 | Night and Day, 1919 | - |
| 117. Walt Whitman | 5 | Leaves of Grass, 1855 | *Poems, 24787* |
| 118. Walter de la Mare | 10 | The Return, 1910 | *Collected Poems, 15765* |
| 119. Washington Irving | 17 | The Legend of Sleepy Hollow, 1820 | - |
| 120. Wilkie Collins | 32 | Hide and Seek, 1854 | - |
| 121. William Blake | 3 | Songs of Innocence, 1789 | *Poems, 4439* |
| 122. William Butler Yeats | 24 | In the Seven Woods, 1903 | *Poems, 23325* |
| 123. William Dean Howells | 84 | Annie Kilburn, 1888 | *Poems, 13554* |
| 124. William Ewart Gladstone | 1 | On Books and the Housing of Them, 1890 | - |
| 125. William Henry Hudson | 13 | The Purple Land, 1885 | - |
| 126. William J Long | 8 | Ways of Wood Folk, 1899 | - |
| 127. William M Thackeray | 30 | Barry Lyndon, 1844 | *Ballads, 20521* |
| 128. William Penn | 2 | A Brief Account of the Rise and Progress of the People Called Quakers, 1698 | - |
| 129. William Shakespeare | 38 | Macbeth, 1623 | *Sonnets, 8721* |
| 130. William Somerset Maugham | 13 | Of Human Bondage, 1915 | - |
| 131. William Wordsworth | 7 | I Wandered Lonely As A Cloud, 1807 | *The Poetical Works, 116683* |
| 132. Winston Churchill (novelist) | 13 | The Inside of the Cup, 1913 | - |

## Top 20 topics with 20 key word stems

Topic 0: much hath name without juan vain hope thine everi less spirit mine doge blood call form youth art think sinc

Topic 1: spirit power hope cloud human deep mountain wave thine among art beneath star blood natur everi wild form around breath

Topic 2: king knight arthur round queen answer mine lancelot saw lord mother arm call name thine hall among child hath speak

Topic 3: king everi wit power much law art kind fate name princ age find grace foe natur arm show without sinc

Topic 4: thir son hath power hast angel self spirit lord stood much hell art find arm without glori less king father

Topic 5: natur marmaduk hath hope power among hill joy oswald everi side round spirit name child sight wood seen father human

Topic 6: back lord king call mother son stand done work run round follow hold soldier watch fight get road tell mine

Topic 7: marmion lord wild king dougla deep vain war saint name band roderick mountain arm knight foe show blood hill everi

Topic 8: angel spirit back mine child thine slowli think toll speak name stand curs adam round sin breath call strong lip

Topic 9: everi wit find think show call much sinc tell lord pleas name muse without better poet virtu art court learn
Topic 10: call near back think mine stood show saw stand yes ere none much sinc hous yea wait name tree woman
Topic 11: king lord bonni ballad son helen green fell three john child father saw queen set grey tell war back round
Topic 12: wit natur everi pleas art fool virtu name learn sens prais vain muse grace fame pride reason call poet lord
Topic 13: king star tree lord sword hous grow grey stand green saw bird hill stood alfr break deer stone wild fell
Topic 14: green breath lip round everi pain silver kiss gentl tell thine feet deep tree doth wide star spirit wild blue
Topic 15: ballad everi kind think tell peter boy name maid pretti marri captain maiden get much doubt willow call tri plan
Topic 16: natur spirit hath everi truth right hope think doth back find much faith art free round whole set drop poet
Topic 17: citi everi hous think noth bodi river state ship stand back shore other wood women pioneer arm star wait mother
Topic 18: king pupil fool cuchullain find hous fintain noth woman call conal run prais barach put anoth laegair tell bodi think
Topic 19: hope joy mother spirit beneath round maid child name power wild deep form breast father natur cloud youth lord gaze

## A Computing the sonority score

Following previous work (Jacobs & Kinder, 2018; Stenneken et al., 2005) and considering that here we deal with written instead of spoken words, I used a simplified index based on the sonority hierarchy of English phonemes which yields 10 ranks: [a] > [e o] > [i u j w] > [ɾ] > [l] > [m n ŋ] > [z v] > [f θ s] > [b d g] > [p t k]. Each word was assigned a value according to the number of graphemes belonging to the 10 rank sets. To control for word length, the sum of the values was divided by the number of graphemes per word. Thus MEMORY would get a value of 9*2 [e o] + 2*5 [m] + 1*7 [r] + 1*8 [y = /i/ ] = 44/6 = 7.33, whereas SKUNK would get a value of 18/5 = 3.6. The final global sonority score of a poem is simply the mean of all word values in the poem. Of course, this simple additive model is only a first approximation, given the lack of any empirical data that would justify more complex models. Moreover, the fact that identical graphemes can have multiple context-dependent pronunciations in English like the /a/ in 'hAndbAll in the pArk' (Ziegler et al., 1997) is neglected in this first approximation which considers written, not spoken verbal materials.

## B Computing word similarity, valence and arousal

Following upon an early unsupervised learning approach proposed by Turney and Littman (2003) and own previous theory-guided research (Westbury et al., 2014), I computed the lexical features *valence* and *arousal* on the basis of (taxonomy-based) semantic associations of a target word with a set of *labels*, i.e. key words assumed to be prototypical for a certain affect, e.g. positive valence. The procedure for computing valence and arousal –implemented as a python script– was as follows. The script compared every target word with every word in the NLTK wordnet/WN database and computed the pairwise similarities (WNsim in equation 1 below, based on WN's path-similarity metric), and summed and averaged them for each target word and for each text.

(1) mean[WNsim(*word*, label_1pos) + ... + WNsim(*word*, label_1Npos)]

label_1pos and label_Npos are the first and last terms, respectively, in either the valence or arousal lists given below.

The hitrates (i.e., overlap between words in the WN database and the present target words) were 80% for the Joyce poem and 77% for Eliot's.

**Label words for computation of positive and negative valence, as well as arousal (for details, see Westbury et al., 2014, Table 2, row 2)**

pos = ['contentment','happiness','pleasure','pride','relief','satisfaction','surprise']

neg = ['disgust','embarrassment','fear','sadness','shame']

aro = ['amusement','anger','contempt','contentment','disgust','embarrassment','excitement','fear','happiness','interest','pleasure','relief','sadness','satisfaction']